\documentclass[11pt,a4paper]{article}
\usepackage[hyperref]{acl2021}
\usepackage{times}
\usepackage{latexsym}
\usepackage{adjustbox}
\usepackage{multirow}
\usepackage{caption}
\usepackage{commath,amsmath, amsfonts, amssymb}
\usepackage{url}
\usepackage{comment}

\usepackage{microtype}

\aclfinalcopy 

\title{A Unified Framework for Emotion Identification and Generation in Dialogues}

\author{Avinash Madasu\thanks{$^*$Both authors contributed equally to this research.} \thanks{$^+$Work done as an intern at IIT Patna.} \unskip\enspace{\rm}\enspace Mauajama Firdaus$^*$ \unskip\enspace{\rm}\enspace Asif Ekbal \\
 Department of Computer Science and Engineering\\
Indian Institute of Technology Patna, India\\ 
{\tt avinashmadasu17@gmail.com};
  {\tt \{mauajama.pcs16,asif}\}{\tt @iitp.ac.in}
}

\begin{document}
\maketitle
\begin{abstract}
Social chatbots have gained immense popularity, and their appeal lies not just in their capacity to respond to the diverse requests from users, but also in the ability to develop an emotional connection with users. To further develop and promote social chatbots, we need to concentrate on increasing user interaction and take into account both the intellectual and emotional quotient in the conversational agents. In this paper, we propose a multi-task framework that jointly identifies the emotion of a given dialogue and generates response in accordance to the identified emotion. We employ a \text{BERT} based network for creating an empathetic system and use a mixed objective function that trains the end-to-end network with both the classification and generation loss. Experimental results show that our proposed framework outperforms current state-of-the-art models. 
\end{abstract}

\section{Introduction}
One of the significant challenges of artificial intelligence (AI) is to endow the machine with the ability to interact in natural language with humans. The personal assistants in our mobile devices invariably assist in our day-to-day lives by answering a wide range of queries. Such assistants act as social agents that take care of the various activities of their users. Besides reacting passively to user requests, they also proactively anticipate user needs and provide in-time assistance, such as reminding of an upcoming event or suggesting a useful service without receiving explicit user requests \cite{sarikaya2017technology}. The daunting task for these agents is that they have to work well in open domain scenarios as people learn to rely on them to effectively maintain their works and lives efficiently.

Building social chatbots to tackle the emotional needs is indeed of great benefit to our society.
The primary objective of these chatbots is not inherently to answer all the users' questions, but rather to be a virtual companion of the users. Therefore, it is essential to empower the conversational agents with the ability to perceive and express emotions to make them capable of interacting with the user at the human level. These agents help enhance user satisfaction \cite{prendinger2005using}, while reducing the breakdowns in conversations \cite{martinovski2003breakdown} and providing user retention. Hence, dialogue systems capable of generating replies while considering the user's emotional state is the most desirable advancement in Artificial Intelligence (AI). 
Previously, researchers have focused upon classifying user emotions \cite{poria2019meld,chauhan2019context} in conversations. For building an intelligent agent, sheer understanding of emotions is insufficient; hence several works \cite{song2019generating,colombo2019affect} have concentrated in inducing emotions into the dialogue system. Most of these existing research have focused on generating emotionally aware \cite{rashkin2019towards,lin2019caire} or emotionally controlled responses \cite{zhou2018emotional,huang2018automatic}. 

In our current work, we focus on creating an end-to-end emotional response generation system that is capable of identifying the emotions and use the emotional information simultaneously for generating empathetic and emotionally coherent responses. The key contributions of our current work are three-fold: (i) We propose a joint model that is able to simultaneously identify the emotions from the utterances and incorporate the emotional information for generation; (ii) We design a multi-task hierarchical framework with a mixed objective function that jointly optimize the classification and generation loss; (iii) Experimental analysis shows that the proposed multi-task framework is better in generating empathetic responses as opposed to single task emotion generation networks.

\section{Related Work}
In \cite{welivita2020fine}, the authors proposed a novel large-scale emotional dialogue dataset, consisting of 1M dialogues and annotated with 32 emotions and 9 empathetic response intents using a \text{BERT}-based fine-grained dialogue emotion classifier. Most of the early research on emotion classification was performed upon textual datasets mostly taken from twitter \cite{colneric2018emotion,ghosal2018contextual}. The authors in \cite{chen2018emotionlines}, propose a multi-party conversational dataset for emotions. Lately, emotional text generation has gained immense popularity \cite{huang2018automatic,li2018syntactically,lin2019caire,li2017dailydialog,ghosh2017affect,rashkin2019towards}. 
In \cite{zhou2018emotional}, an emotional chatting machine (ECM) was built based on seq2seq framework for generating emotional responses. 
Recently, a lexicon-based attention framework was employed to generate responses with a specific emotion \cite{song2019generating}. Emotional embedding, affective sampling and regularizer were employed to generate the affect driven dialogues in \cite{colombo2019affect}. The authors employed curriculum dual learning \cite{shen2020cdl} for emotion controllable response generation. In \cite{asghar2018affective,lubis2018eliciting,zhong2019affect,li2019empgan}, the authors proposed an end-to-end neural framework that captures the emotional state of the user for generating empathetic responses. Our present research differs from these as we propose and address a novel task of generating responses and simultaneously identifying the emotions in a multi-task framework. 

\section{Problem Definition}
In this paper, we address the problem of identifying emotion from a user utterance and generate empathetic responses accordingly. In this setting, emotion remains the same throughout the conversation. Let $U_p = u_{p,1}, u_{p,2}, ... , u_{p,j}$ be the set of utterances in a conversation and $E_{p}$ denotes the emotion for the conversation. We aim to identify $E_{p}$ through our emotion classification sub-network and use this emotion information to generate responses to the user utterances $U_{p}$. Understanding emotion initially is very crucial to empathetic response generation and cannot be treated independently. Hence, we aim to propose an end-to-end model capable to identifying emotions and utilizes this emotion information for generation.
\begin{figure}[!h]
    \centering
    \includegraphics[scale=0.5]{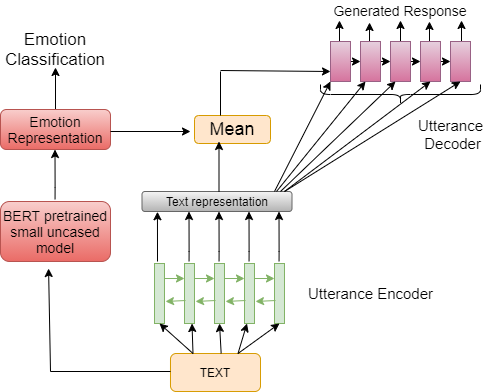}
    \caption{Architecture of the proposed model}
    \label{arch}
\end{figure}

\section{Proposed Model}
In this section, we present the proposed model. It has two sub-networks: Emotion classification sub-network and Generation sub-network.
\subsection{Emotion classification sub-network}
The input to this sub-network is a user utterance and the output is predicted emotion for this utterance. We employed a pre-trained \text{BERT} \cite{devlin-etal-2019-bert} small uncased model. Instead of taking $[CLS]$ token from just the final layer, we learn a weighted representation on $[CLS]$ tokens from all the layers.
\begin{equation}\label{emorep}
\begin{aligned}
{h_{cls}^{l}} &= \text{BERT}(U_{p})    l \in \{1,\dotsc, 12\}\\
C_{E} &= \sum_{i=1}^{l} {\alpha_{t,i}} {h_{cls}^{l}}  \\
\alpha_{t,i} &= softmax({{h_{cls}^{l}}^T}{W_g}{h_{cls, t}}) \\
P(\tilde{y_{em}} \mid y_{em}) &= softmax({W_{E}} {C}_{E}),
\end{aligned}
\end{equation}
$y_{em}$ is the true emotion labels and $\tilde{y_{em}}$ is the predicted emotion label.
The emotion classifier is trained by minimizing the negative log-likelihood
\begin{equation}
    \mathcal{L}_{Emo} = -\sum_{em=1}^{N} y_{em} \log \tilde{y_{em}}
\end{equation}

\subsection{Generation sub-network}
\textbf{Utterance Encoder} For a given utterance $U_p$, we employ a bidirectional Gated Recurrent Units (Bi-GRU) \cite{cho2014learning} to encode each word $u_{p,j}$, where $j \in (1,2,3,.....n)$ having $d$-dimensional embedding vectors into the hidden representation $h_{U_k,i}$. We concatenate the last hidden representation from both the unidirectional GRUs to form the final hidden representation of a given utterance as follows:  
\begin{equation}
    h_{U_p,i}^{en} = [GRU ({u_{p,j}},\overrightarrow{h_{U_p,i-1}}), GRU ({u_{p,j}},\overleftarrow{h_{U_p,i-1}})]
\end{equation}
\textbf{Utterance Decoder}
For decoding the response to a given user utterance, we build a GRU which takes encoder last hidden state as the initial hidden state and words generated previously. To integrate emotion information, we compute the mean of emotion representation obtained using $C_{E}$ and the encoder's last hidden state to initialize as decoder's initial state. We use the input feeding decoding along with the attention \cite{luong2015effective} mechanism for enhancing the performance of the model. Using the decoder state $h_{d,t}^{dec}$ as the query vector, we apply self-attention on the hidden representation of the utterance encoder. The decoder state and the encoder vector are concatenated and used to calculate a final distribution of the probability over the output tokens.
\begin{equation}
\begin{split}
h_{d,{t-1}} = Mean(C_{E}, h_{c, t}^{en})
\\
    h_{d,t}^{dec} = GRU_d(y_{k,t-1}, h_{d,{t-1}})
\\
    c_t = \sum_{i=1}^{k} {\alpha_{t,i}} {h_{c,p}^{ctx}}, 
\\
\alpha_{t,i} = softmax({{h_{c,p}^{en}}^T}{W_f}{h_{d,t}})
\\
 \tilde{h}_t = tanh(W_{\tilde{h}}{[{h_{d,t}};c_t]})
\\
P(\tilde{y_{t}} \mid y^*_{t}) = softmax({W_V}{\tilde{h}_t})
\end{split}
\end{equation}
where, $W_f$, $W_V$ and $W_{\tilde{h}}$ are the trainable weight matrices. $y^*_{t}$, $\tilde{y_{t}}$ are the ground truth and generated words at each time-step respectively. 
We employ the teacher forcing \cite{williams1989learning} algorithm at every decoding step to minimize the negative log-likelihood on the model distribution.
\begin{equation}
\mathcal{L}_{Gen} = -\sum_{t=1}^{m} \log p(y^*_t | y^*_1, \ldots, y^*_{t-1})
\label{eq:ml-loss}
\end{equation}

\subsection{Joint Training (JT)}
To train an end-to-end model, we jointly optimize the emotion classification loss and the generation loss. The final loss of the entire model is:
\begin{equation}
    \mathcal{L}_{JT} = \mathcal L_{Emo} + \mathcal L_{Gen}
\end{equation}

\section{Dataset and Experimentation}
\textbf{Dataset}
We conduct our experiment on the EmpatheticDialogues \cite{rashkin2019towards} dataset which consist of 25k open-domain conversations grounded in emotional situations and provides 32 emotion classes. We used the train, test splits provided by the authors. \\
\textbf{Implementation Details}
The hidden size of utterance encoder Bi-GRU is 768 and the hidden size of utterance decoder GRU is 768. A dropout of 0.3 is applied on both the utterance encoder and decoder layers. A dropout of 0.1 is applied on the weighted emotion representation obtained from \ref{emorep} just before the final softmax layer. All the models are trained with a batch size of 32 for 10 epochs. AdamW is used as the optimizer with a learning rate of 0.0001. The maximum sentence length used is 80.

\section{Baseline Methods and Metrics}
We compare our model to the below SoTA models. \\
\textbf{Pretrained:} \\
It is a transformer model trained on 1.7 billion REDDIT conversations \cite{lin2019caire}. \\
\textbf{Fine-Tuned:} \\
The above pre-trained model is fine-tuned on Emotion Dialogue Dataset \cite{lin2019caire}. \\
\textbf{MULTITASK:} \\
It is an emotion classifier trained using a linear layer on the top of a transformer \cite{lin2019caire}. \\
\textbf{EmoPrepend-1:} \\
A top-1 predicted emotion is appended to the beginning of input sentence \cite{lin2019caire}. \\
\textbf{ENSEM-DM:} \\
The representations obtained from transformer encoder and pre-trained emotion classifier are concatenated and then fed to transformer decoder \cite{lin2019caire}. \\
\textbf{CAiRE:} \\
It is the current State-of-the-art model. In this model, a large language model is jointly trained for multiple objectives response language modeling, response prediction, and dialogue emotion detection. It is then fine-tuned on EmphatheticDialogues dataset \cite{lin2019caire}. \\ 
\textbf{Bi-LSTM-Attn (JT):} \\
To signify the importance of a \text{BERT} pre-trained model, we replace \text{BERT} with Bi-LSTM layer with attention applied on the top of it.  \\
\textbf{Metrics:} \\
We use AVG BLEU, the average of BLEU-1, BLEU-2, BLEU-3, BLEU-4 \cite{Papineni2002BleuAM} and emotion F1 for comparison. As the dataset is unbalanced, we didn't use emotion accuracy for models' comparison. We used balanced emotion labels during manual evaluation and hence we included emotion accuracy.
\begin{table}[h]
\centering
\caption{Automatic evaluation results of emotion identification and generation. Here: JT-joint training }\label{result}
\begin{tabular}{c|cc}
\hline
\textbf{Models} & \textbf{\shortstack{AVG \\ BLEU}} & \textbf{\shortstack{EMO \\ F1}} \\ \hline
 Pretrained \cite{lin2019caire} & 5.01&- \\
Fine-Tuned \cite{lin2019caire} & 6.27&- \\
 MULTITASK \cite{lin2019caire} &5.42 &- \\
 EmoPrepend-1 \cite{lin2019caire} &4.36 &- \\
 ENSEM-DM \cite{lin2019caire} &6.83 &- \\
 CAiRE \cite{lin2019caire} &7.03 &- \\ \hline
 Bi-LSTM (JT) & 6.84 & 8.22 \\
Bi-LSTM + Attn (JT) & 6.13 &19.89 \\
\textit{\textbf{\text{BERT} (JT)}}  & \textbf{7.71} & \textbf{25.2} \\ \hline
\end{tabular}
\end{table}

\section{Result and Analysis}
In this section, we present the results of our proposed framework. We compute average BLEU scores  for the model response, comparing against the gold label (the actual response). From the Table \ref{result}, it is evident that our proposed framework performs significantly better in comparison to the existing baselines\footnote{all the results are statistically significant. We perform statistical significance t-test \cite{welch1947generalization}, and it is conducted at 5\% (0.05) significance level}. As compared to CAiRE, we see an improvement of 0.68 in average BLEU score. Emotion identification is crucial for relevant empathetic responses and the generated responses should be in accordance to sentiment expressed by the user. Joint training helps in learning better emotion representations thereby generating contextually coherent, interactive, engaging and empathetic responses. The emotion classification results reported in Table \ref{result}, demonstrates effectiveness of joint training. From table \ref{result}, it is visible that the F1-score of \text{BERT} based classifier is better in comparison to the joint training Bi-LSTM and Bi-LSTM + Attn with a gain in performance of 17\% and 6\% respectively.

\begin{table}
\centering
\caption{Human evaluation results of emotion identification and generation. Here: Flu-Fluency; Cons-Consistency; EmoAcc-Emotion Accuracy }\label{human}
\begin{tabular}{c|ccc}
\hline
\textbf{Model} & \textbf{Flu} & \textbf{Cons} & \textbf{EmoAcc} \\ \hline
Bi-LSTM (JT)  & 3.82 & 3.73 & 48\% \\
Bi-LSTM + Attn (JT) & 3.93 & 3.82 & 55\% \\ \hline
\textit{\textbf{\text{BERT} (JT)}} &\textbf{4.11} & \textbf{3.96} & \textbf{69\%} \\ \hline
\end{tabular}
\end{table}

We recruit six annotators (in a similar manner as \cite{shang2015neural,tian2019learning}) from a third party company, having high-level language skills. In Table \ref{human} we present the results of human evaluation. We sampled 250 responses per model for evaluation with the utterance and the conversational history provided for generation. First, we evaluate the quality of the response on two conventional criteria: \textit{Fluency} and \textit{Consistency}. These are rated on a five-scale, where 1, 3, 5 indicate unacceptable, moderate, and excellent performance, respectively, while 2 and 4 are used for unsure.
Secondly, we evaluate the emotion quotient of a response in terms of \textit{Emotion Accuracy} metric
that measures whether the emotion induced in the response is in accordance with the predicted emotion information and the dialogue history. 
Here, 0 indicates irrelevant or contradictory, and 1 indicates consistent with the predicted emotion and dialogue context. We compute Fleiss' kappa \cite{fleiss1971measuring} to measure inter-rater consistency. The Fleiss' kappa for Fluency and Consistency is 0.53 and 0.49, indicating moderate agreement. For Emotion Accuracy, we get 0.65 as the kappa scores indicating substantial agreement. From the table, it is evident that our proposed framework performs better for all the human evaluation metrics. The responses generated are fluent, consistent to the dialogue history and the emotion quotient of the generated response is higher in comparison to the baselines. Also, the joint training mechanism proves to be efficient for simultaneously identifying the emotions and using the emotional information for generation.

\section{Conclusion and Future Work}
In our current work, we propose the task of jointly identifying the emotions in dialogues and use the emotional information for generating empathetic responses. For our proposed task, we design a \text{BERT}-based multi-task framework, that simultaneously identifies the emotion of the speaker and generate the response in accordance to the situation, conversational history and the predicted emotions. Experimental analysis on the EmpatheticDialogues dataset shows that the proposed framework achieved State-of-the-art results. In future, we look forward to designing sophisticated joint mechanisms  to enhance empathetic response generation.

\bibliographystyle{acl_natbib}
\bibliography{anthology,acl2021}


\end{document}